\documentclass[conference, anonymous]{IEEEtran}
\IEEEoverridecommandlockouts

\usepackage{cite}
\usepackage{amsmath,amssymb,amsfonts}
\IEEEoverridecommandlockouts

\usepackage{graphicx}
\usepackage{xcolor}
\usepackage{subcaption}
\usepackage{float}

\usepackage{tabularx}
\usepackage{multirow}
\usepackage[table,xcdraw]{xcolor}

\usepackage{algorithm}
\usepackage{algpseudocode}
\definecolor{lightgreen}{rgb}{0.3, 0.7, 0.3}
\algrenewcommand\algorithmiccomment[1]{\hfill\textcolor{lightgreen}{// #1}}

\usepackage[hidelinks]{hyperref}
\hypersetup{colorlinks=true, citecolor=green}

\title{VesSAM: Efficient Multi-Prompting for Segmenting Complex Vessel}

\begin{document}

\author{
\centering
\begin{minipage}{\textwidth}
\centering
\setlength{\baselineskip}{1.1em} % 控制行距
\sloppy % 允许换行在逗号处

Suzhong Fu$^{1,2}$, %
Rui Sun$^{1,2}$, %
Xuan Ding$^{1,2}$, %
Jingqi Dong$^{1,2}$, %
Yiming Yang$^{1,2}$, %
Yao Zhu$^{3}$, %
Min Chang Jordan Ren$^{4}$, %
Delin Deng$^{5}$, %
Angelica Aviles-Rivero$^{6}$, %
Shuguang Cui$^{1,2}$, %
Zhen Li$^{1,2,*}$\\[4pt]

$^{1}$FNii-Shenzhen, CUHK-Shenzhen, Shenzhen, China\\
$^{2}$School of Science and Engineering, CUHK-Shenzhen, Shenzhen, China\\
$^{3}$Zhejiang University, Hangzhou, China 
$^{4}$Boston University, United States\\
$^{5}$Vanderbilt University, United States 
$^{6}$Tsinghua University, Beijing, China\\[4pt]
\end{minipage}

}

\maketitle

\begin{abstract}
Precise vessel segmentation is vital for clinical applications such as diagnosis and surgical planning but remains challenging due to thin, branching geometries and low texture contrast.Although foundation models such as the Segment Anything Model (SAM) show strong performance in general segmentation tasks, they remain suboptimal for vascular structures.
In this work, we present \textbf{VesSAM}, a powerful and efficient framework tailored for 2D vessel segmentation. VesSAM integrates three core modules: a convolutional adapter that enhances local texture features, a multi-prompt encoder that fuses anatomical cues via hierarchical cross-attention, and a lightweight mask decoder that reduces jagged artifacts. We also introduce an automated pipeline to generate structured multi-prompt annotations, and curate a diverse benchmark dataset spanning 8 datasets across 5 imaging modalities.
Extensive experiments show that VesSAM surpasses state-of-the-art PEFT-based SAM variants by over \textbf{10\% Dice} and \textbf{13\% IoU}, while maintaining competitive accuracy to fully fine-tuned methods with far fewer parameters. VesSAM also generalizes well to out-of-distribution (OoD) settings, outperforming all baselines in average OoD Dice and IoU.

\end{abstract}


\section{Introduction}

Accurate segmentation of vascular structures is vital for clinical applications such as disease diagnosis, surgical planning, and treatment monitoring. Unlike solid organs, vessels display thin, elongated geometries with complex branching patterns, posing unique challenges for automatic segmentation. Although deep learning-based methods have achieved remarkable success in vessel segmentation~\cite{fu2023robust,yao2025high,yang2023robust,isensee2021nnu}, they often rely on dense pixel-wise annotations, which are expensive and time-consuming to obtain in clinical workflows.

\renewcommand{\thefootnote}{}
\footnotetext{*Corresponding author: lizhen@cuhk.edu.cn}
\renewcommand{\thefootnote}{\arabic{footnote}} % 恢复正常编号

Recently, foundation models~\cite{touvron2023llama,achiam2023gpt,glm2024chatglm,liu2024grounding} such as the Segment Anything Model (SAM)~\cite{kirillov2023segment} have enabled prompt-based, domain-adaptable segmentation. In the medical domain, SAM-based extensions~\cite{ma2024segment,zhang2023customized,cheng2023sam,wang2023sam} have demonstrated promising results for organ and tumor segmentation. These models use generic prompts—such as boxes or sparse points—to guide segmentation, reducing annotation burden. However, their performance on vessel segmentation remains limited.

\begin{figure}[t!]
    \centering
    \includegraphics[width=0.5\textwidth]{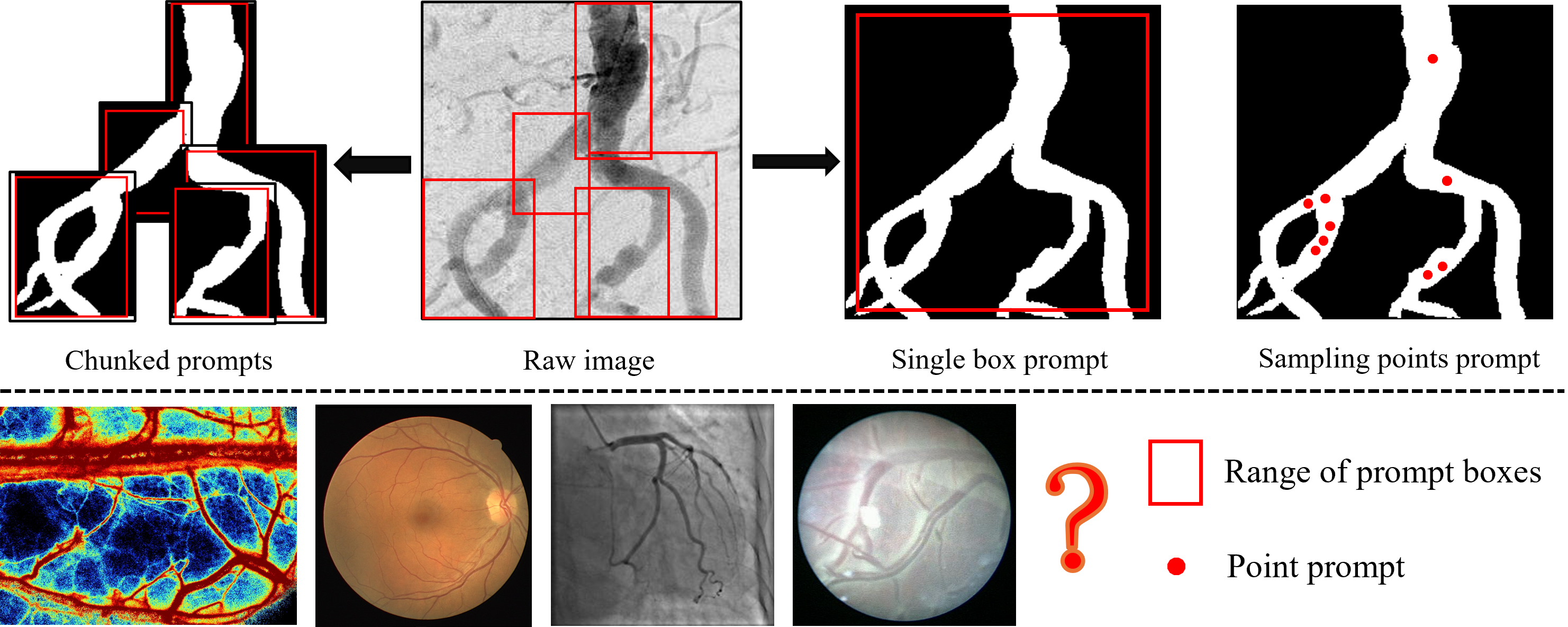}
    \caption{
    Motivating examples showing the limitations of conventional prompt strategies in vessel segmentation. 
    % Simple structures can be handled by boxes or random points, but these methods fail on complex, dense vascular networks—highlighting the need for more informative, structure-aware prompting.
    }
    \label{Figure1}
\end{figure}

As illustrated in Fig.~\ref{Figure1}, conventional prompting strategies perform adequately in simple vascular cases but degrade in complex ones. Full-image boxes offer little spatial constraint, while random point sampling within irregular masks introduces bias and uneven supervision. These methods struggle with fine, dense vascular networks characterized by intricate bifurcations and low contrast, where traditional prompts fail to capture detailed topology or maintain vessel continuity.

Beyond prompting, model architecture also constrains performance. Existing SAM-based frameworks rely heavily on ViT backbones. While transformers excel at modeling long-range dependencies, their patch-based tokenization can neglect local texture continuity—especially in non-convex, sparse structures like vessels. In organ segmentation, patch-wise attention aligns well with intra-mask continuity, but in vessel segmentation, masks often span multiple disconnected or narrow regions, undermining global modeling alone. Therefore, a successful vessel segmentation system must integrate both global context and local detail through texture-sensitive and topologically meaningful representations.

% VesSAM introduces: (1) a convolutional adapter to enhance ViT with localized texture sensitivity and efficient fine-tuning; (2) a multi-prompt encoder to integrate skeletons, bifurcation points, and midpoints via hierarchical cross-attention and graph reasoning; and (3) a lightweight mask decoder for smooth boundary refinement through coarse-to-fine upsampling.
To address these issues, we propose VesSAM, a structure-aware and parameter-efficient segmentation framework tailored for vascular imaging. 
We construct a multi-modality vessel dataset with automatically generated prompt annotations to support structure-aware training.
Code is publicly available.\footnote{\url{https://github.com/VersaceSu/VesSAM}}

\vspace{0.5em}
\noindent \textbf{Our key contributions are as follows:}

\begin{itemize}
    \item We propose \textbf{VesSAM}, a segmentation framework that enhances ViT-based encoders with localized texture modeling and anatomical prompt fusion, significantly improving accuracy on non-convex vascular structures.
    \item We design a \textbf{multi-prompt encoder} that unifies sparse (e.g., bifurcation points) and dense (e.g., skeletons, masks) anatomical cues using hierarchical cross-attention and graph-based topology reasoning.
    % \item We introduce a \textbf{convolutional adapter} that improves fine-detail representation while enabling parameter-efficient tuning via LoRA-style isolation.
    \item We develop an automated prompt generation pipeline and release a \textbf{benchmark dataset spanning eight vascular datasets across five imaging modalities}, supporting scalable, high-fidelity supervision.
\end{itemize}

% \begin{figure}[h!]
% \centering
% \includegraphics[width=0.49\textwidth]{figure/Figure2.png}
% \caption{Examples from different datasets. Top row: raw vessel images; middle row: skeletons with bifurcation points (red) and midpoints (green); bottom row: final segmentation masks with prompt annotations.}
% \label{Figure2}
% \end{figure}

\section{Experiments Setting}

% Most SAM-related medical segmentation methods focus on large convex organs, where sparse prompts (e.g., bounding boxes or random points) are often sufficient. However, these strategies are inadequate for segmenting \textbf{non-convex and topologically complex structures} such as blood vessels, where global spatial prompts lack the fine-grained details granularity and connectivity prompts.

% To address this limitation, 
\subsection{Datasets}
We construct a comprehensive vessel benchmark by aggregating \textbf{eight datasets across five imaging modalities}, each annotated with precise vessel delineations. Specifically, the dataset includes pelvic-iliac artery angiograms (Aorta)~\cite{zohranyan2024dr}, coronary XCAD~\cite{ma2021self}, retinal datasets (ARIA~\cite{Far2008enhan}, DRIVE~\cite{staal2004ridge}, HRF~\cite{1282003}, IOSTAR~\cite{zhang2016robust}), placental vessel images (PSVFM)~\cite{bano2020deep}, and laser speckle contrast imaging (LSCI)~\cite{fu2023robust}.

% All samples are provided with expert-validated ground truth annotations.

\textbf{Multi-Prompt Strategy.}  
We employ a multi-prompt strategy that designs three anatomically grounded prompt types to guide segmentation:
 \textbf{Bifurcation points:} minimal keypoints to resolve topological ambiguity;
 \textbf{Segment midpoints:} orientation-aware anchors that reduce redundancy;
 \textbf{Skeleton maps:} structural templates that preserve global vessel continuity.
To systematically extract these prompts, we develop an automatic prompt generation algorithm,
% , shown in Algorithm~\ref{algorithm1}
which processes vessel masks to generate informative prompt sets. 

\subsection{Experiments Setting}

\textbf{Baseline Methods.}
To ensure a fair comparison with existing methods, we benchmark VesSAM against five representative baselines. Among them, SAM-Med2D~\cite{cheng2023sam}, MedSAM~\cite{ma2024segment}, and SAMed~\cite{zhang2023customized} represent parameter-efficient fine-tuning (PEFT) extensions of the Segment Anything Model (SAM) and are adapted to the vascular domain with lightweight adapters or LoRA modules. We also include two fully fine-tuned state-of-the-art methods: nnUNet~\cite{isensee2021nnu} and TransUNet~\cite{chen2021transunet}, which serve as strong supervised baselines in medical image segmentation. All methods are fine-tuned on the same training splits and evaluated under identical settings.

\textbf{Ablation Studies.}
To analyze the contribution of each component in VesSAM, we conducted comprehensive ablation studies. We evaluated the impact of different prompt combinations by testing six configurations: using only bifurcation points, only segment midpoints, only skeleton maps, bifurcation+midpoints, bifurcation+skeletons, and all three combined. These experiments were performed under  $512 \times 512$ input resolutions, as shown in Fig.~\ref{Figure5}.

\begin{figure}[!ht]
\centering
\includegraphics[width=0.5\textwidth]{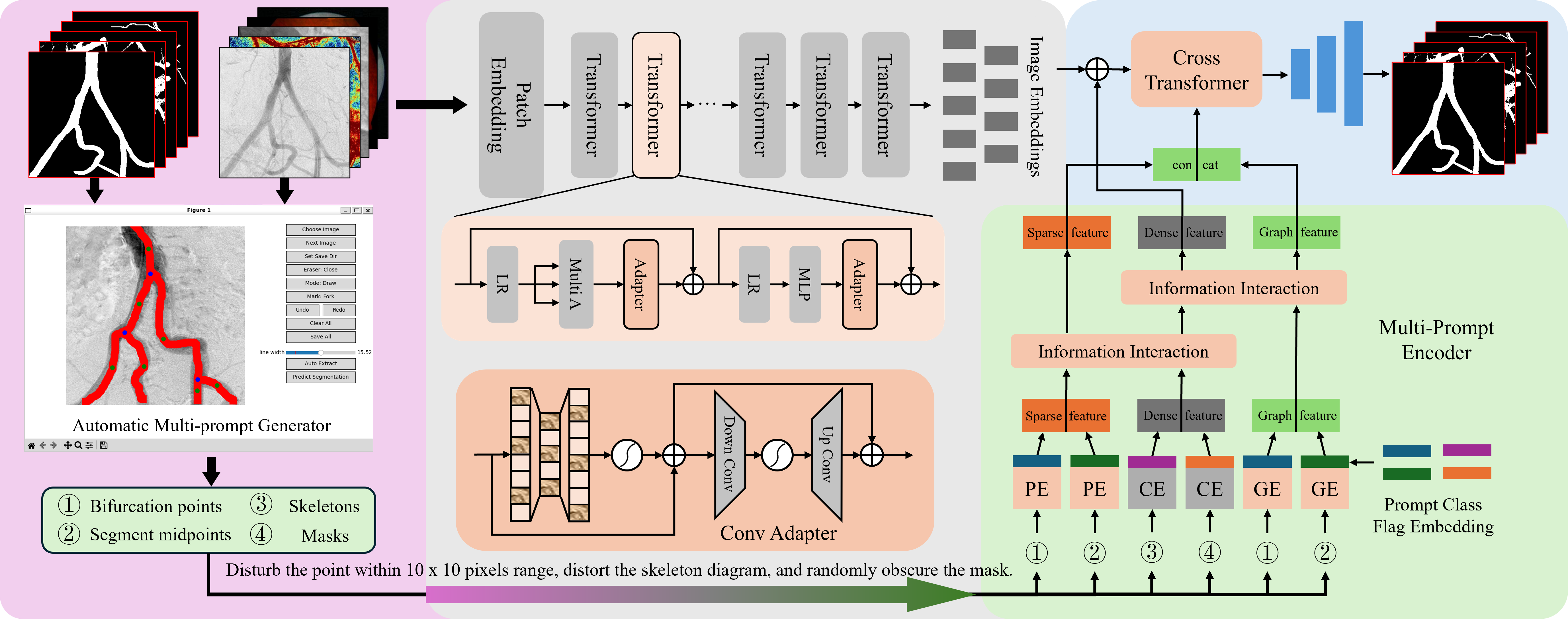}
\caption{
Overview of the proposed \textbf{VesSAM} framework. }

\label{Figure3}
\end{figure}

\section{Method}

VesSAM addresses vessel segmentation challenges with a prompt-aware and texture-sensitive design. As shown in Fig.~\ref{Figure3}, it consists of three components: a convolution-enhanced image encoder, a multi-prompt encoder integrating diverse anatomical cues, and a lightweight mask decoder for fine-grained segmentation. 

% An automated multi-prompt generator further extracts structural prompts—such as skeletons, bifurcations, and midpoints—from vessel masks to guide the process.

\begin{table*}[!t]
    \caption{Segmentation Performance on 512*512 (H) resolution dataset. The best results are in bold  with brown background, and the second-best are underlined.}
    \centering
    \begin{tabular*}{\linewidth}{@{\extracolsep{\fill}}lcccccccccccc@{}}
    \hline
    \multicolumn{1}{c}{} & \multicolumn{6}{c}{\textbf{PEFT Methods}} & \multicolumn{6}{c}{\textbf{Full Fine-tuning Methods}} \\
    Dataset & \multicolumn{2}{c}{VesSAM} & \multicolumn{2}{c}{SAM-Med2D} & \multicolumn{2}{c}{SAMed} & \multicolumn{2}{c}{MedSAM} & \multicolumn{2}{c}{nnUNet} & \multicolumn{2}{c}{TransUnet} \\
    Metric & IoU & Dice & IoU & Dice & IoU & Dice & IoU & Dice & IoU & Dice & IoU & Dice \\
    \hline
LSCI & 77.10 & 87.01 & 61.49 & 75.73 & 55.76 & 70.97 & 74.15 & 84.88 & \cellcolor[HTML]{D2B48C}\textbf{82.66} & \cellcolor[HTML]{D2B48C}\textbf{90.43} & \underline{81.26} & \underline{89.16} \\
Placenta & \cellcolor[HTML]{D2B48C}\textbf{74.38} & \cellcolor[HTML]{D2B48C}\textbf{84.61} & 56.35 & 70.58 & 71.74 & 82.55 & \underline{73.86} & \underline{84.22} & 69.50 & 81.03 & 68.10 & 79.67 \\
Retinal & \cellcolor[HTML]{D2B48C}\textbf{70.10} & \cellcolor[HTML]{D2B48C}\textbf{82.32} & 24.29 & 38.91 & 41.57 & 58.50 & \underline{66.03} & \underline{79.16} & 60.86 & 75.58 & 59.05 & 74.12 \\
Aorta & 92.33 & 95.99 & 84.02 & 91.26 & 93.25 & 96.50 & 93.06 & 96.39 & \cellcolor[HTML]{D2B48C}\textbf{93.82} & \cellcolor[HTML]{D2B48C}\textbf{96.78} & \underline{93.49} & \underline{96.61} \\
XCAD & \cellcolor[HTML]{D2B48C}\textbf{85.54} & \cellcolor[HTML]{D2B48C}\textbf{92.14} & 47.05 & 63.62 & 70.74 & 82.78 & \underline{82.42} & \underline{89.72} & 71.42 & 83.23 & 68.82 & 81.41 \\
ALL & \cellcolor[HTML]{D2B48C}\textbf{78.66} & \cellcolor[HTML]{D2B48C}\textbf{87.36} & 51.93 & 64.45 & 54.01 & 62.52 & 76.39 & 86.08 & \underline{78.14} & \underline{86.70} & 77.19 & 86.21 \\
Average(H) & \cellcolor[HTML]{D2B48C}\textbf{79.89} & \cellcolor[HTML]{D2B48C}\textbf{88.41} & 54.64 & 68.02 & 66.61 & 78.26 & \underline{77.90} & \underline{86.87} & 75.65 & 85.41 & 74.14 & 84.19 \\
\hline
    \end{tabular*}
    \label{Table512_id}
\end{table*}

\begin{table*}[!t]
    \caption{Performance under OoD setting. The best results are in bold  with brown background, and the second-best are underlined. Labeled in the table for testing, and the other four for training.}
\centering
    \begin{tabular*}{\linewidth}{@{\extracolsep{\fill}}lcccccccccccc@{}}
    \hline
    \multicolumn{1}{c}{} & \multicolumn{6}{c}{\textbf{PEFT Methods}} & \multicolumn{6}{c}{\textbf{Full Fine-tuning Methods}} \\
    Dataset & \multicolumn{2}{c}{VesSAM} & \multicolumn{2}{c}{SAM-Med2D} & \multicolumn{2}{c}{SAMed} & \multicolumn{2}{c}{MedSAM} & \multicolumn{2}{c}{uuUNet} & \multicolumn{2}{c}{TransUnet} \\
    Metric & IoU & Dice & IoU & Dice & IoU & Dice & IoU & Dice & IoU & Dice & IoU & Dice \\
    \hline
    LSCI & 24.27 & 38.02 & 8.25 & 14.48 & 26.28 & 40.40 & 24.77 & 39.05 & \underline{27.73} & \underline{40.99} & \cellcolor[HTML]{D2B48C}\textbf{38.14} & \cellcolor[HTML]{D2B48C}\textbf{51.87} \\
    retinal & \cellcolor[HTML]{D2B48C}\textbf{50.25} & \cellcolor[HTML]{D2B48C}\textbf{66.76} & 19.65 & 32.71 & 16.27 & 27.09 & 10.30 & 17.89 & 24.77 & 38.45 & \underline{32.30} & \underline{48.45} \\
    Placenta & \cellcolor[HTML]{D2B48C}\textbf{23.94} & \cellcolor[HTML]{D2B48C}\textbf{36.99} & 10.93 & 17.59 & \underline{22.20} & \underline{31.47} & 0.87 & 1.67 & 7.53 & 12.13 & 12.32 & 18.85 \\
    Aorta & \underline{80.55} & \underline{89.10} & 53.44 & 68.87 & \cellcolor[HTML]{D2B48C}\textbf{82.07} & \cellcolor[HTML]{D2B48C}\textbf{89.90} & 34.25 & 49.01 & 61.31 & 75.38 & 72.74 & 84.02 \\
    XCAD & \cellcolor[HTML]{D2B48C}\textbf{61.55} & \cellcolor[HTML]{D2B48C}\textbf{76.16} & 41.39 & 58.16 & \underline{54.81} & \underline{70.34} & 30.86 & 46.89 & 38.22 & 53.81 & 48.06 & 64.33 \\
    Average & \cellcolor[HTML]{D2B48C}\textbf{48.11} & \cellcolor[HTML]{D2B48C}\textbf{61.40} & 26.73 & 38.36 &40.33 & 51.84 & 20.21 & 30.90 & 31.91 & 44.15 &  \underline{40.71} & \underline{53.50} \\
    \hline
    \end{tabular*}
    \label{Table512_Ood}
\end{table*}

% We also validate the effectiveness of the proposed cross-feature fusion mechanism, we tested four configurations: no fusion, fusion between sparse and dense features (SD), fusion between dense and graph features (DG), and combined SD+DG fusion. Results across five single-modality datasets and the aggregated dataset are illustrated in Fig.~\ref{Figure6}(a). 

% Lastly, we ablated the proposed random prompt dropout strategy by comparing two setups on the full $256 \times 256$ dataset: one with random omission of prompts during training, and one with all prompts always present. As shown in Fig.~\ref{Figure6}(b), incorporating stochastic dropout yields more robust performance in prompt-sparse environments.

\textbf{Image Encoder with Convolutional Adapter}.
Although SAM employs a ViT-based~\cite{dosovitskiy2020image} encoder to capture long-range dependencies, it lacks sensitivity to the local continuity and texture crucial for thin, non-convex vessels. Unlike organs with coherent and convex structures, vessels extend across spatially separated patches, requiring stronger modeling of local cues beyond patch boundaries.

To address this, we introduce a convolutional adapter that complements the ViT encoder. It employs depth-wise separable convolutions for efficient channel attention and a lightweight spatial attention branch using a downsampling–upsampling sequence to capture coarse patterns and refine details. By combining channel and spatial attention, the adapter enhances ViT’s global modeling with local structural awareness, improving performance on fine vessel structures with minimal parameter overhead.

\textbf{Multi-Prompt Encoder}.
VesSAM leverages rich anatomical cues through multi-type prompts that include sparse point sets (bifurcations and midpoints), dense maps (skeleton and mask), and topological relationships. These prompts are processed through a dedicated encoder designed to preserve their distinct spatial and semantic roles while allowing effective feature fusion.

For sparse prompts, each point set is first embedded with a learnable coordinate encoder and tagged with a prompt-type indicator (e.g., bifurcation or midpoint). The resulting representations are concatenated and passed through a lightweight convolutional block to yield sparse features.
For dense prompts, both the skeleton map and mask are independently encoded through shallow convolutional encoders, producing dense features.
To capture global vessel topology, we additionally construct a graph using the bifurcation and midpoint prompts as nodes, and apply graph convolution to encode their relational structure, resulting in graph features with topological knowledge.

These three sources of information—sparse features ($\mathbf{SF}$), dense features ($\mathbf{DF}$), and graph features ($\mathbf{GF}$)—are then integrated via a two-stage cross-attention mechanism. In the first stage, sparse and dense features are fused.
\begin{equation}
\mathbf{SF}', \mathbf{DF}' = \text{CrossAttention}(\mathbf{SF}, \mathbf{DF})
\end{equation}
In the second stage, the updated dense features interact with the graph features to yield topology-enhanced representations:
\begin{equation}
\mathbf{GF}', \mathbf{DF}'' = \text{CrossAttention}(\mathbf{DF}', \mathbf{GF})
\end{equation}

The outputs $\mathbf{SF}'$, $\mathbf{DF}''$, and $\mathbf{GF}'$ are then forwarded to the mask decoder for final prediction.

\textbf{Lightweight Mask Decoder}.
The decoder module integrates features from the image encoder and the multi-prompt encoder to produce the final segmentation mask. First, global image tokens extracted by the ViT encoder are concatenated with the prompt features $\mathbf{SF}'$, $\mathbf{DF}''$, and $\mathbf{GF}'$. Second, the combined sequences processed by a transformer-based attention module for cross-modal fusion. 

% Unlike SAM’s decoder, which relies on transposed convolutions and is susceptible to producing jagged artifacts—especially in texture-rich regions—our design adopts a progressive upsampling strategy. The decoder reconstructs the mask in a coarse-to-fine manner, applying anisotropic Gaussian smoothing at each upsampling stage to mitigate aliasing effects and preserve anatomical continuity. Finally, a lightweight convolutional head outputs the predicted segmentation mask.
% Through its modular and prompt-aware design, VesSAM effectively integrates local texture, sparse anatomical landmarks, and global vessel topology. 

\begin{figure}[!t]
    \centering
    \includegraphics[width=0.5\textwidth]{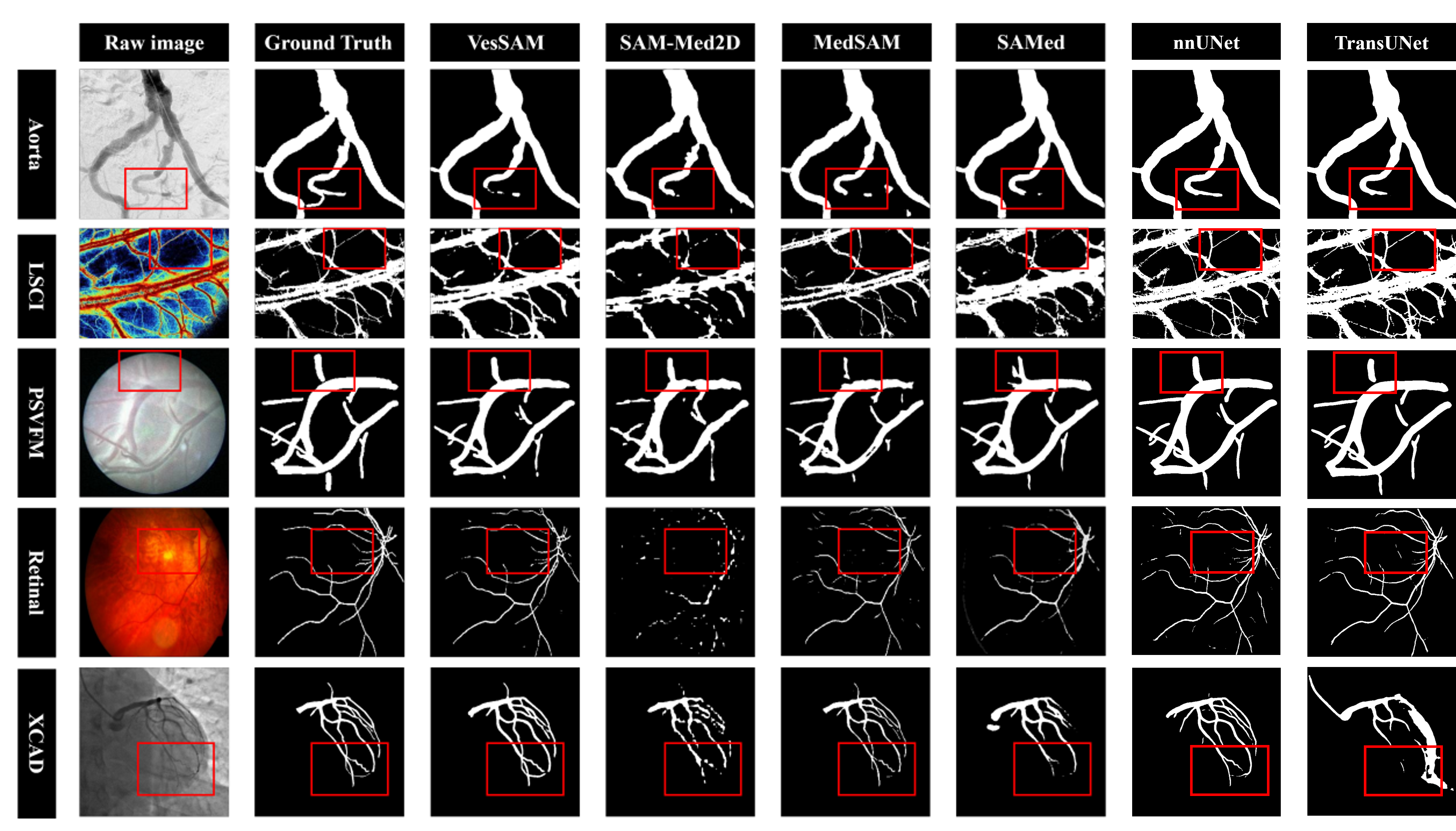}
    \caption{Visualization of segmentation results from different methods.}
    % , presentation of five types of modal data, methods include PEFT fine-tuned VesSAM, SAM-Med2D, SAMed, and fully fine-tuned MedSAM, uuUNet, and TransUNet.
    \label{Figure4}
\end{figure}

\section{Experiments Results}

\begin{figure}[!t]
    \centering
    \includegraphics[width=0.5\textwidth]{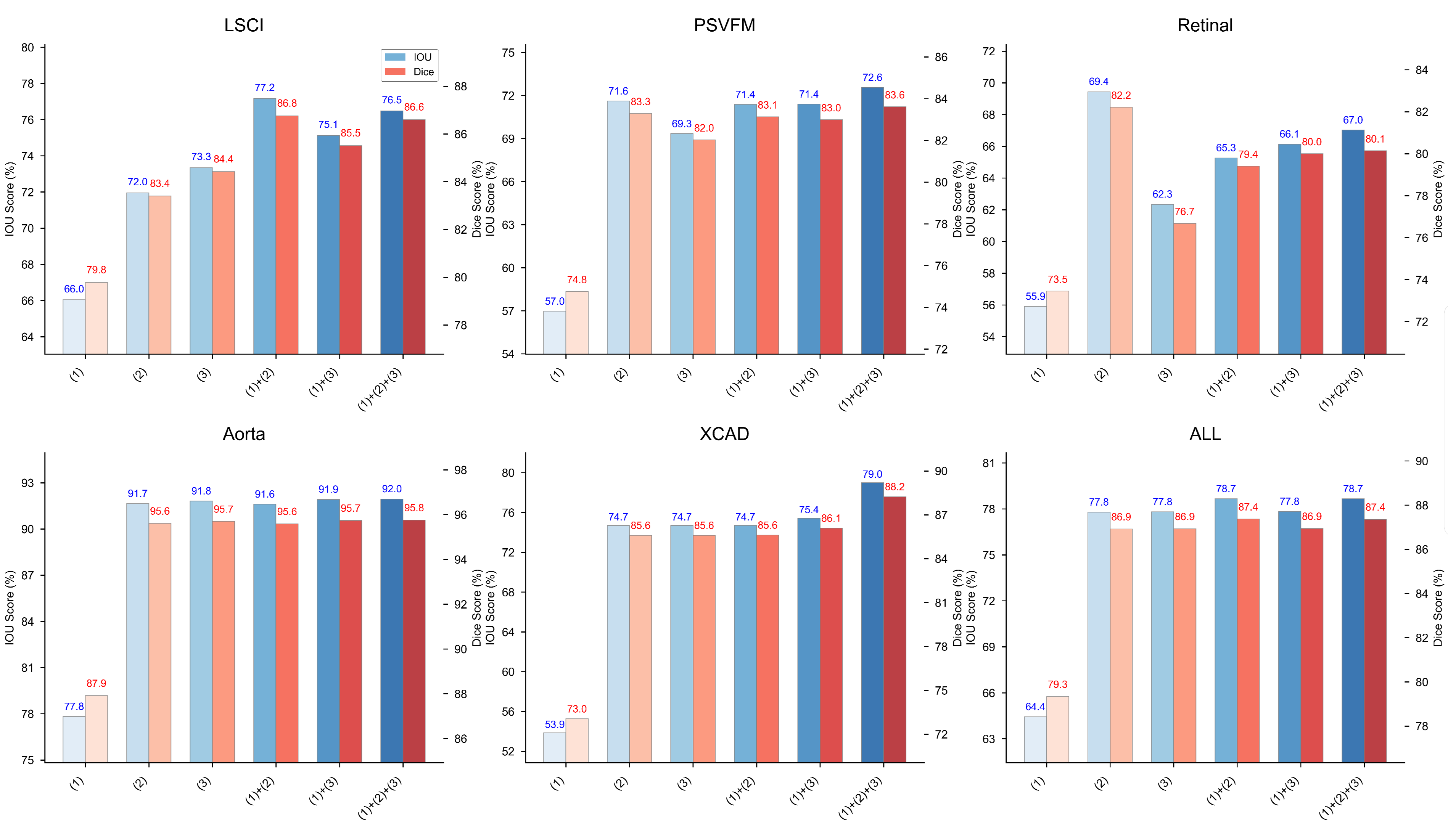}
    \caption{Ablation on different prompt combinations. 
    % Using multiple complementary prompts (branch points, midpoints, and skeletons) yields the highest Dice and IoU across datasets.
    }
    \label{Figure5}
\end{figure}

\textbf{Performance under In-Distribution (ID) Settings.}
We evaluated at  $512 \times 512$ resolution datasets, VesSAM exhibits even stronger performance. As shown in Table~\ref{Table512_id}, the model not only outperforms SAMed by more than 10\% in Dice and 13\% in IoU, but also surpasses nnUNet by 3\% in Dice score. The advantage becomes especially pronounced on datasets such as LSCI and Retina, which require fine-grained boundary preservation and vascular continuity. The multi-prompt mechanism is particularly effective in suppressing noise, preserving thin structures, and improving segmentation robustness. VesSAM also yields strong results on PSVFM and XCAD datasets, highlighting its ability to handle blurry boundaries and small-scale vessel features. 

% It is worth noting that MedSAM resizes all inputs to $1024 \times 1024$, potentially offering it a resolution-based advantage in this comparison.

\textbf{Performance under Out-of-Distribution (OoD) Settings.}
To evaluate cross-domain generalization, we conduct experiments under the out-of-distribution (OoD) setting by holding out one modality as the test domain while training on the remaining ones. As shown in Table~\ref{Table512_Ood}, VesSAM achieves an average Dice score of 61.4\% and an IoU of 48.11\%, marking an improvement of 9.56\% and 9.78\% over SAMed, respectively. Compared to other methods, VesSAM exhibits greater resilience on challenging datasets such as Retinal and XCAD, where complex vascular structures and terminal branches require detailed supervision. In contrast, SAM-Med2D, SAMed, nnUNet, and MedSAM suffer substantial performance drops on these datasets, particularly in scenarios involving small vessels and boundary ambiguity. These results underscore the importance of structure-aware prompting in achieving robust generalization across heterogeneous imaging domains.

\textbf{Visualization Analysis.}
Qualitative comparisons further highlight VesSAM’s strengths. As illustrated in Fig.~\ref{Figure4}, across five modalities, VesSAM consistently produces smoother and more anatomically accurate vessel masks. For instance, in the Aorta dataset, where noise artifacts resemble vessel endpoints, MedSAM is easily prone to false positives, while VesSAM leverages keypoint prompts to preserve structure and reject irrelevant features. TransUNet and nnUNet, although competitive, often miss thin branches or introduce artifacts in low-contrast regions. VesSAM also outperforms others on the LSCI dataset through effective suppression of background noise while preserving fine vascular continuity. On PSVFM and XCAD, where boundary quality is compromised, VesSAM benefits from stronger prompt encoding and convolutional refinement, yielding sharper and more complete segmentation. In contrast to the jagged artifacts observed in SAM-Med2D outputs, VesSAM’s mask decoder produces more consistent and artifact-free results.

\section*{Acknowledgment}
The work was supported in part by NSFC with Grant No. 12326610\&62573371, by Guangdong S\&T Programme with Grant No. 2024B0101030002, by the Basic Research Project No. HZQB-KCZYZ-2021067 of Hetao ShenzhenHK S\&T Cooperation Zone, by the Shenzhen General
Program No. JCYJ2022053014360001, by the Shenzhen Outstanding Talents Training Fund 202002, by the Guangdong Research Project No. 2017ZT07X152 and No. 2019CX01X104, by the Guangdong Provincial Key Laboratory of Future Networks of Intelligence (Grant No. 2022B1212010001), by the Guangdong Provincial Key Laboratory of BigData Computing CHUKShenzhen, by the NSFC 61931024\&62293482, by the Key Area R\&D Program of Guangdong Province with grant No. 2018B030338001, by the Shenzhen Key Laboratory of Big Data and Artificial Intelligence (Grant No. SYSPG20241211173853027), by China Association for Science and Technology Youth Care Program, by the Shenzhen-Hong Kong Joint Funding No.SGDX20211123112401002, and by Tencent \& Huawei Open Fund.

% \vspace{12pt}
% \color{red}
% IEEE conference templates contain guidance text for composing and formatting conference papers. Please ensure that all template text is removed from your conference paper prior to submission to the conference. Failure to remove the template text from your paper may result in your paper not being published.

\end{document}